\def\year{2018}\relax
\documentclass[letterpaper]{article} 
\usepackage{aaai18}  
\usepackage{times}  
\usepackage{helvet}  
\usepackage{courier}  
\usepackage{url}  
\usepackage{graphicx}  
\usepackage{booktabs}
\usepackage{subcaption}
\usepackage{adjustbox}
\usepackage{multirow}
\usepackage{lineno}
\usepackage{amsmath}

\newcommand{\citet}[1]
{\citeauthor{#1}~\shortcite{#1}}
\newcommand{\citep}{\cite}

\begin{document}
%
\title{Who Said What: Modeling Individual Labelers Improves Classification}

\author{Melody Y. Guan\thanks{Work done as a member of the Google Brain Residency program (g.co/brainresidency).}\\
Stanford University\\
450 Serra Mall\\
Stanford, California 94305\\
mguan@stanford.edu
\And
Varun Gulshan, Andrew M. Dai, Geoffrey E. Hinton\\
Google Brain\\
1600 Amphitheatre Pwky\\
Mountain View, California 94043\\
\{varungulshan, adai, geoffhinton\}@google.com
}
\maketitle

\begin{abstract}
Data are often labeled by many different experts with each expert only labeling a small fraction of the data and each data point being labeled by several experts. This reduces the workload on individual experts and also gives a better estimate of the unobserved ground truth. When experts disagree, the standard approaches are to treat the majority opinion as the correct label and to model the correct label as a distribution. These approaches, however, do not make any use of potentially valuable information about which expert produced which label. To make use of this extra information, we propose modeling the experts individually and then learning averaging weights for combining them, possibly in sample-specific ways. This allows us to give more weight to more reliable experts and take advantage of the unique strengths of individual experts at classifying certain types of data. Here we show that our approach leads to improvements in computer-aided diagnosis of diabetic retinopathy. We also show that our method performs better than competing algorithms by \citet{welinder}; \citet{mnih}. Our work offers an innovative approach for dealing with the myriad real-world settings that use expert opinions to define labels for training.
\end{abstract}

\begin{figure}
\centering
\includegraphics[width=0.474\textwidth]{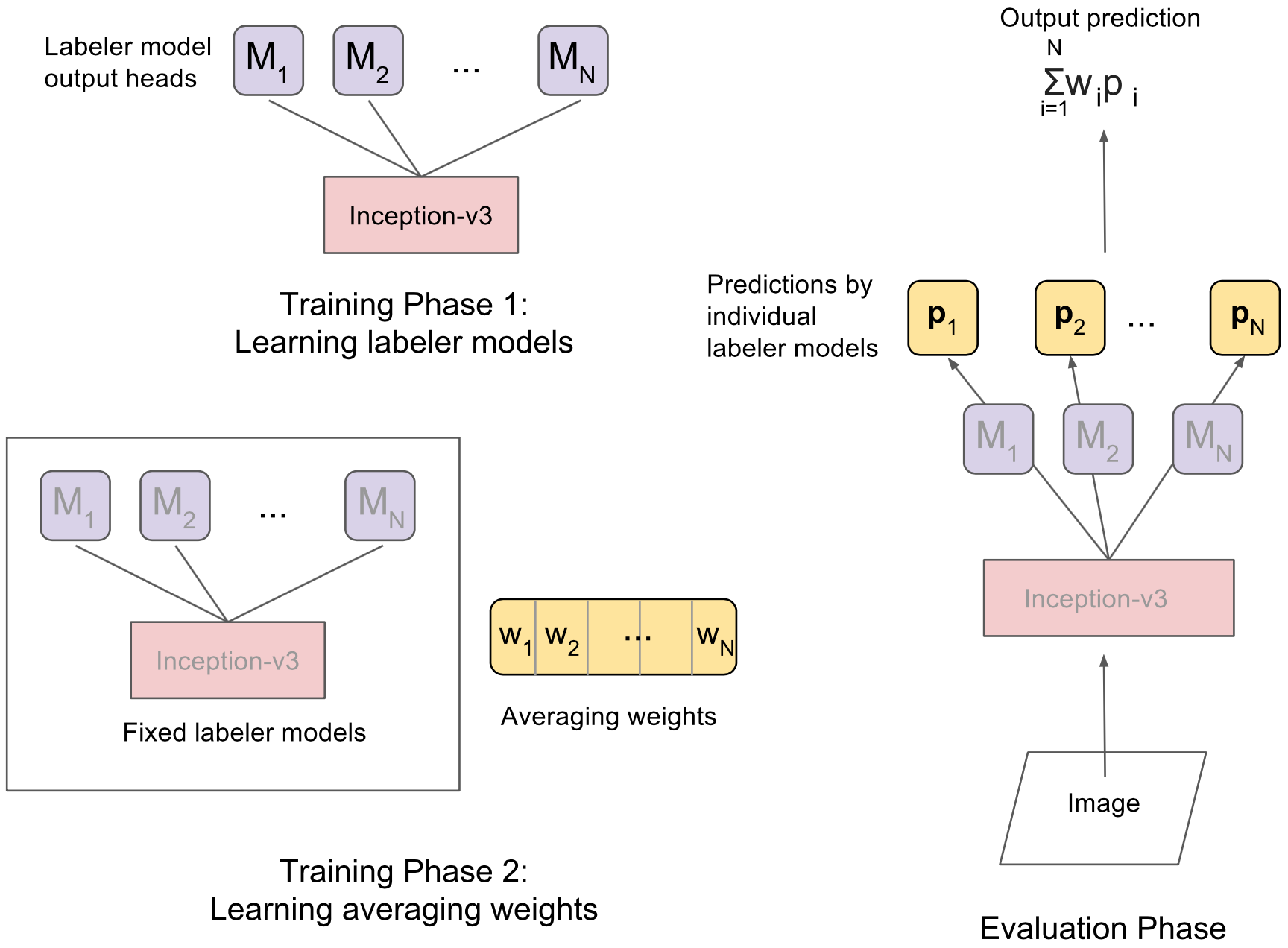}
\caption{\label{fig:flowchart} Overview of proposed procedure to learn classification from multiple noisy annotators.}
\end{figure}

\section{Introduction}
Over the last few years, deep convolutional neural networks have led to rapid improvements in the ability of computers to classify objects in images and they are now comparable with human performance in several domains. As computers get faster and researchers develop even better techniques, neural networks will continue to improve, especially for tasks where it is possible to get a very large number of accurately labeled training examples. In the near future, we can expect neural networks to start serving as alternatives to human experts. We would, in fact, like the neural networks to perform much better than the human experts used to provide the training labels because these training labels are often unreliable as indicated by the poor agreement between different experts ($55.4\%$ for the datasets we consider) or even between an expert and the same expert looking at the same image some time later ($70.7\%$).
\footnote{Inter-grader variability is a well-known issue in many settings in which human interpretation is used as a proxy for ground truth, such as radiology and pathology \cite{elmore,elmore2}.}
Intuitively, we would expect the quality of the training labels to provide an upper bound on the performance of the trained net. In the following section we show that this intuition is incorrect.

Our paper's main contribution is to show that there are significantly better ways to use the opinions of multiple experts than simply treating the consensus of the experts as the correct label or using the experts to define a probability distribution over labels. 

Figure~\ref{fig:flowchart} summarizes our optimal procedure.

\begin{figure}
\centering
\includegraphics[width=0.47\textwidth]{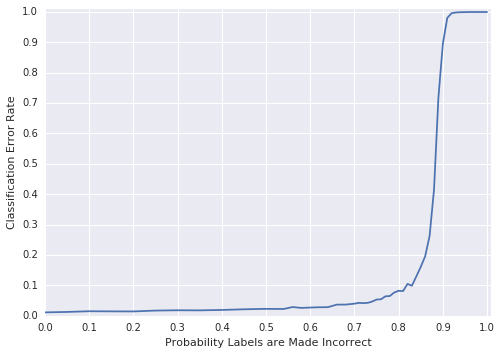}
\caption{\label{fig:catastrophe} Performance of a deep neural net when trained with noisy labels.}
\end{figure}

\subsection{Beating the Teacher}

To demonstrate that a trained neural net can perform far better than its teacher we use the well-known MNIST hand-written digit benchmark for which the true labels are known and we create unreliable training labels by corrupting the true labels. This corruption is performed just once per experiment, before training starts, so the noise introduced by the corruption cannot be averaged away by training on the same example several times. MNIST has 60k training images and 10k test images and the task is to classify each 28$\times$28-pixel image into one of the ten classes. 
For the purposes of this demonstration, we use a very simple neural net: two convolutional layers with 5$\times$5 filters, rectified linear unit (ReLU) activation functions, and 16 and 25 output channels respectively, each followed by a max pooling layer with 2x2 filters; a fully connected hidden layer of 32 ReLUs; and a 10-way softmax layer. We train the net on 50k examples using stochastic gradient descent on mini-batches of size 200 with the Adam optimizer \cite{kingma} and we use the remaining 10k training cases as a validation set for tuning the learning rate and the magnitude of the initial random weights. The best-performing net has a test error rate of 1.01\% when the training labels were all correct. 
If the labels are corrupted by changing each label to one of the other nine classes with a probability of 0.5, the test error rate only rises to 2.29\%. Even if each training label is changed to an incorrect label with probability 0.8 so that the teacher is wrong 80\% of the time, the trained net only gets 8.23\% test error. If the teacher is even less reliable there comes a point at which the neural net fails to ``get the point'' and its error rate rises catastrophically, but this does not happen until the teacher is extremely unreliable as shown in Figure \ref{fig:catastrophe}.

This demonstrates that the performance of a neural net is not limited by the accuracy of its teacher, provided the teacher's errors are random. One obvious question is how many noisily labeled training examples are worth a correctly labeled training example. In Appendix A we show that this question can be answered, at least approximately, by computing the mutual information between label and truth.

\subsection{Making Better Use of Noisy Labels for Diabetic Retinopathy Classification}
We are interested in noisy datasets of medical images where many different doctors have provided labels but each image has only been labeled by a few doctors and most of the doctors have only labeled a fairly small fraction of the images. This paper focuses on datasets of images used for screening diabetic retinopathy because neural networks have recently achieved human-level performance on such images \cite{gulshan} and if we can produce even a relatively small improvement in the state-of-the-art system it will be of great value.

\begin{figure}
\centering
\subcaptionbox{Healthy} {
\includegraphics[width=0.45\linewidth]{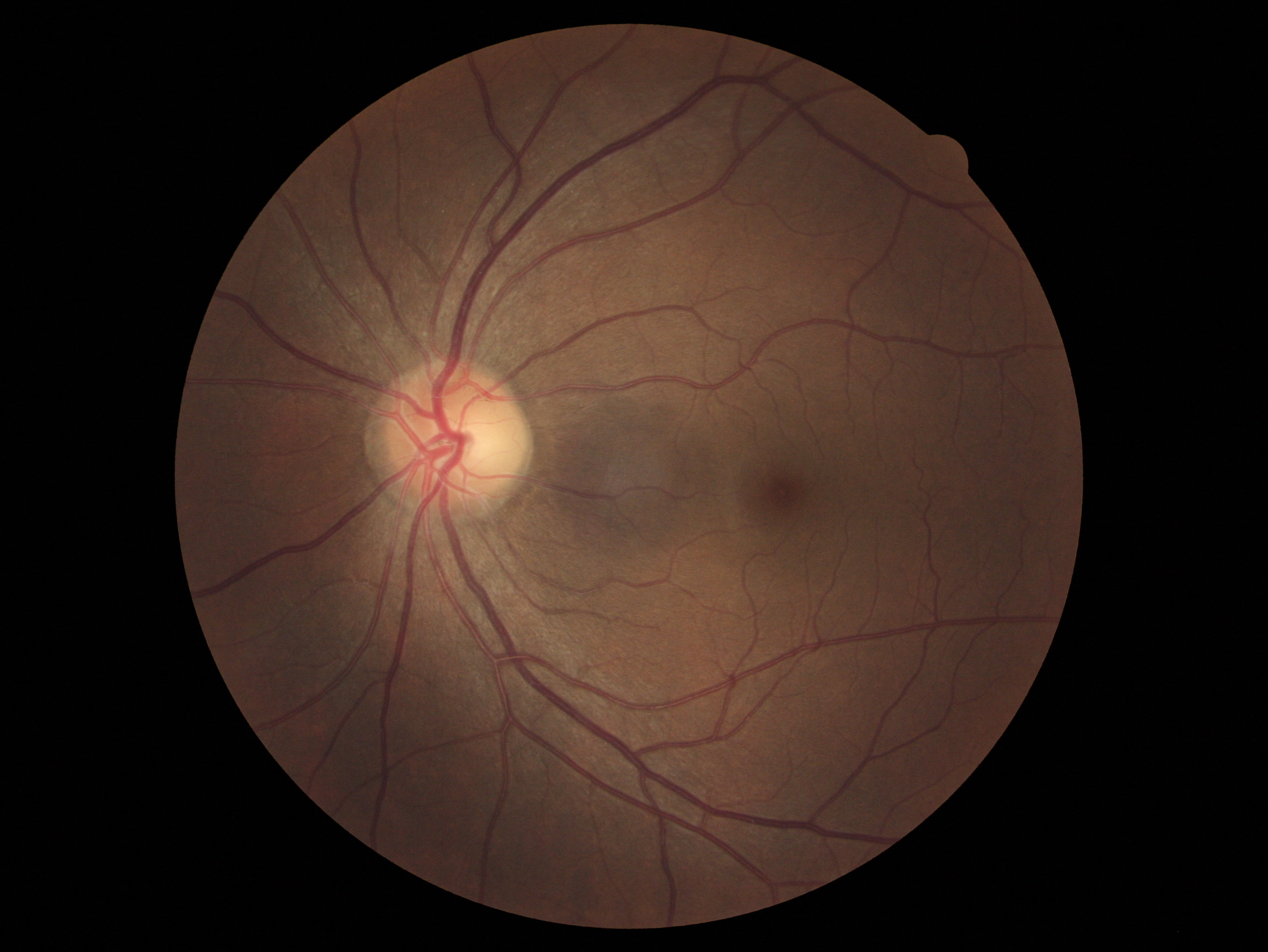}
}
\subcaptionbox{Mild NPDR} {
\includegraphics[width=0.45\linewidth]{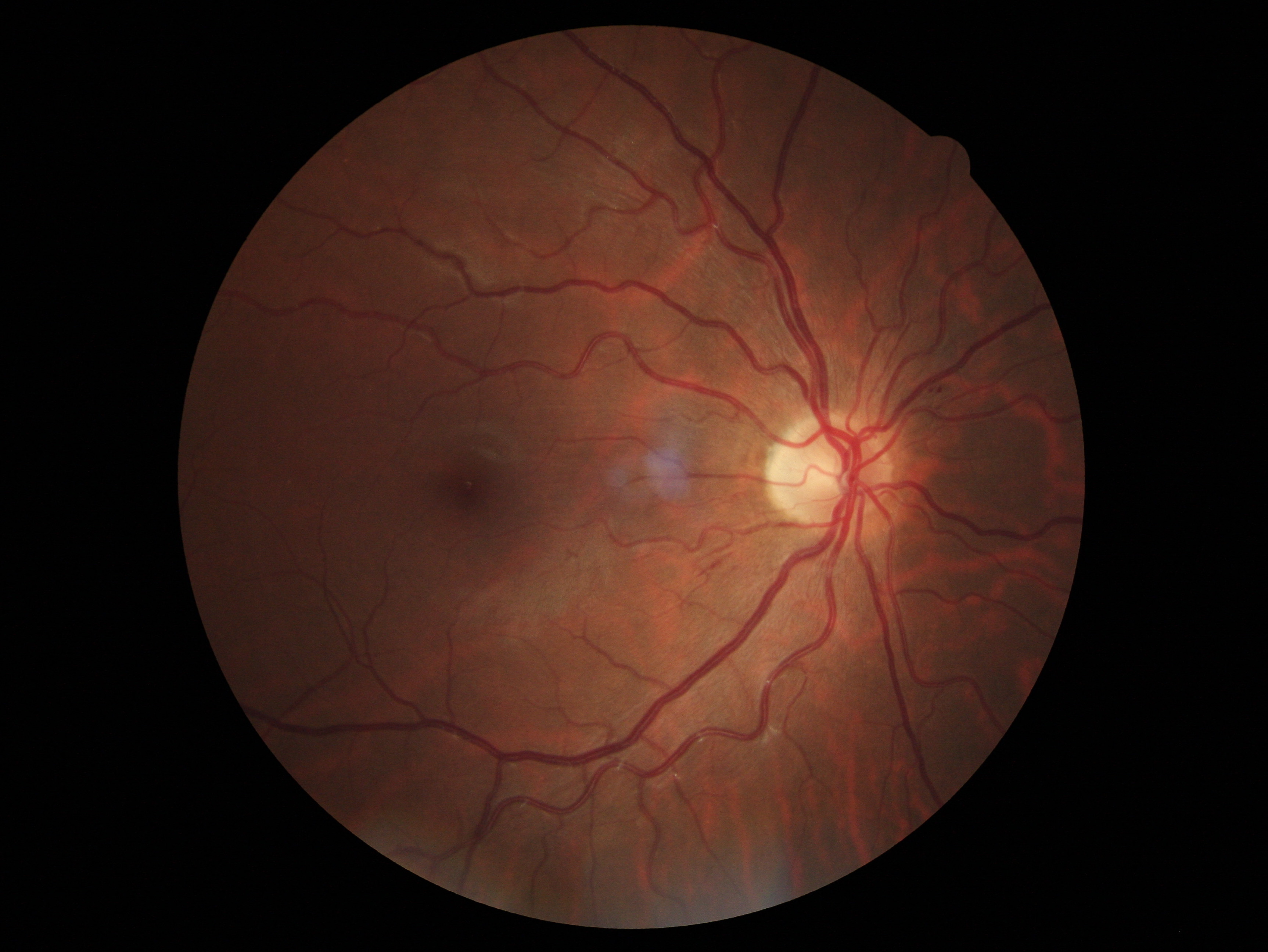}
}
\subcaptionbox{Moderate NPDR} {
\includegraphics[width=0.45\linewidth]{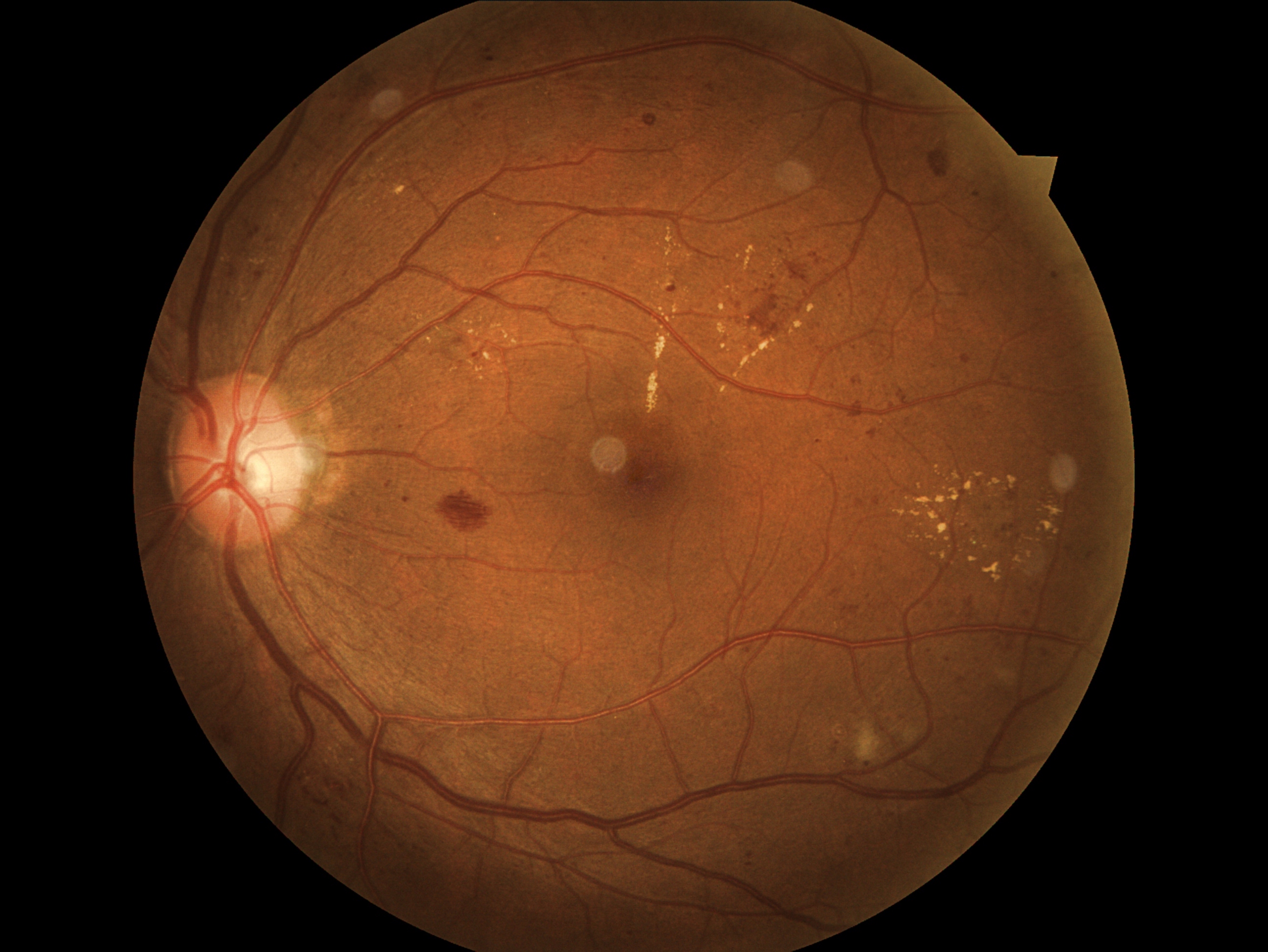}
}
\subcaptionbox{Severe NPDR} {
\includegraphics[width=0.45\linewidth]{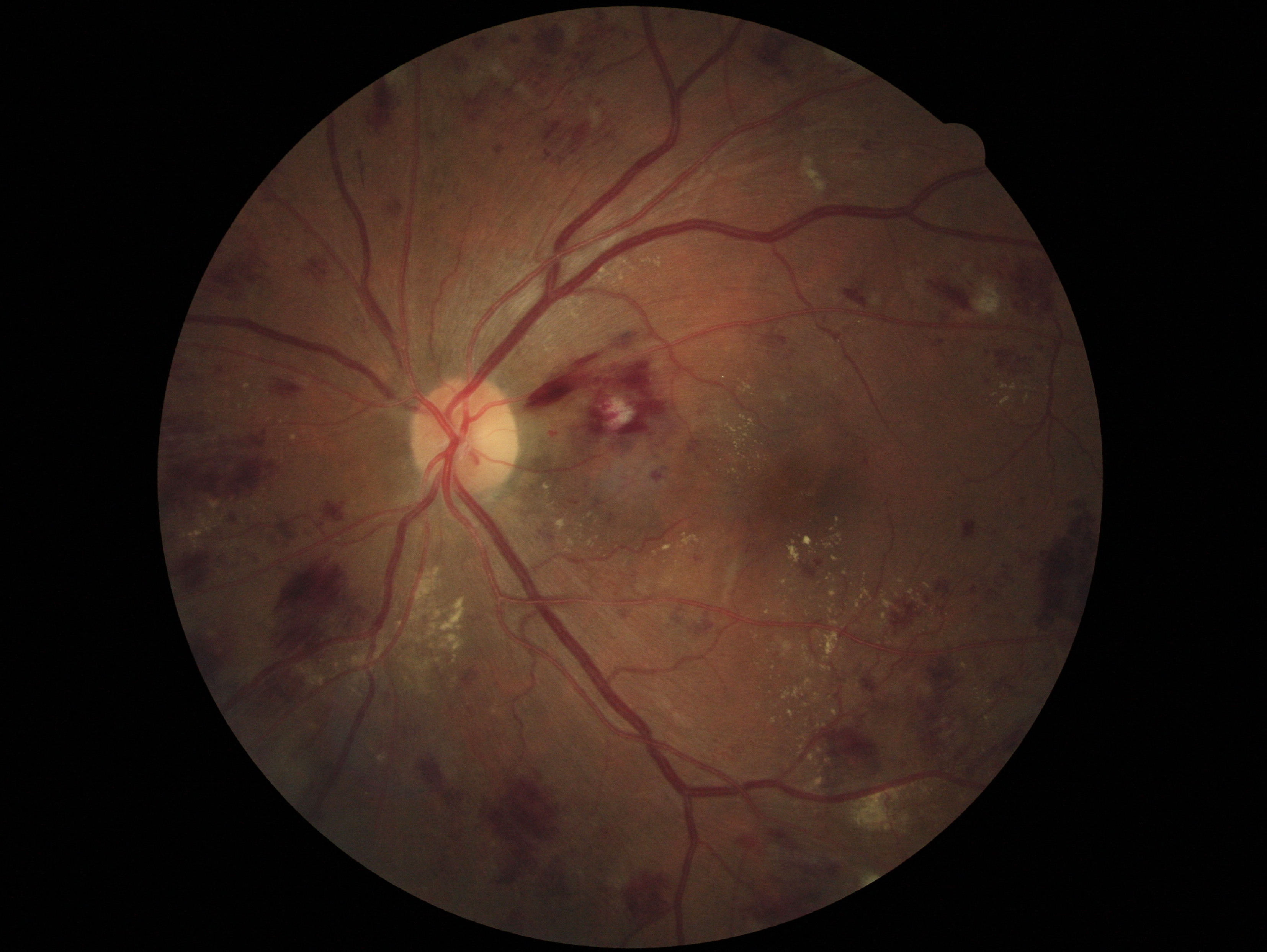}
}
\subcaptionbox{Proliferative DR} {
\includegraphics[width=0.45\linewidth]{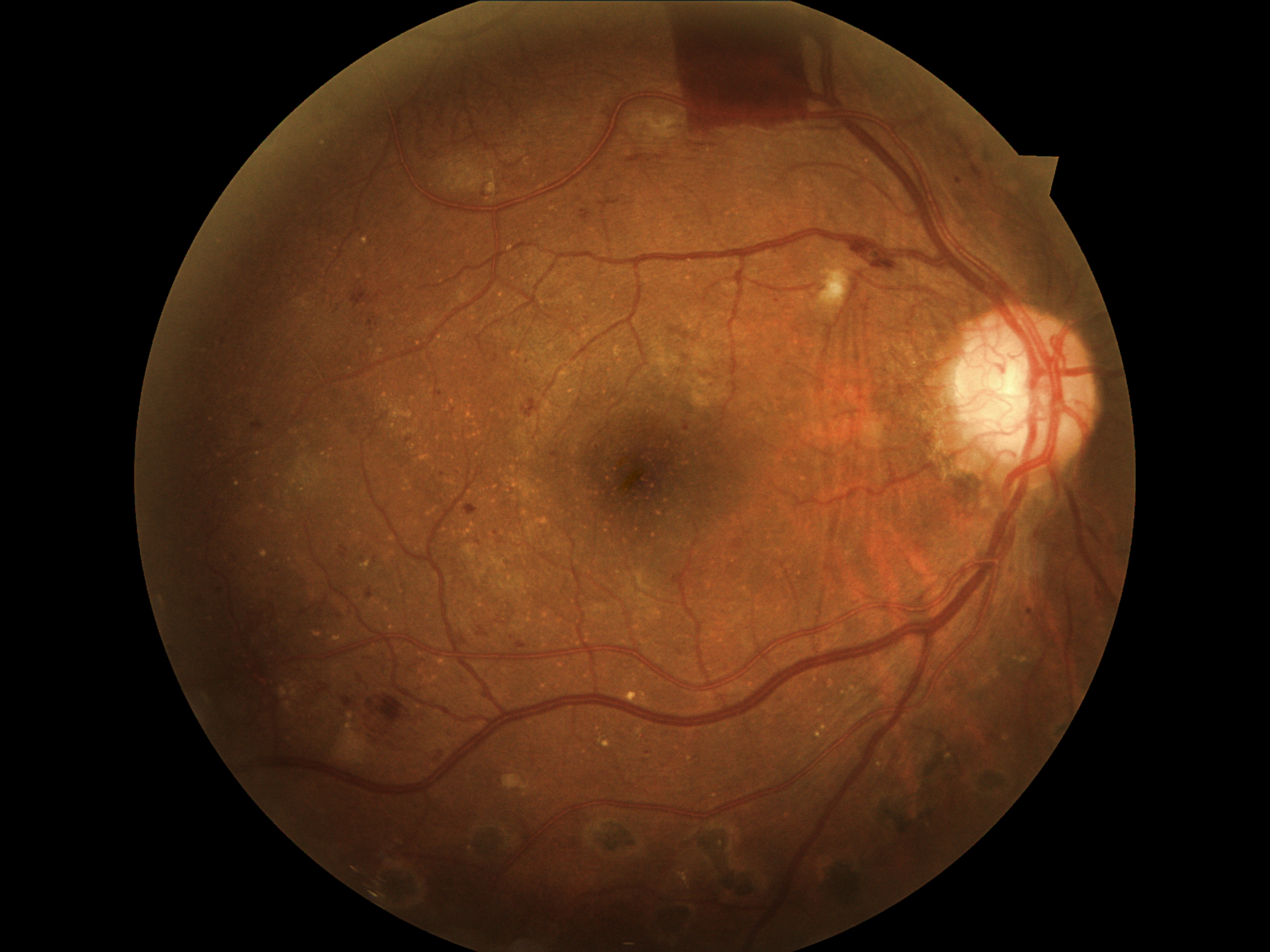}
}
\caption{Sample fundus images from each DR class.}
\label{fig:fundus_images}
\end{figure}

Diabetic retinopathy (DR) is the fastest growing cause of blindness worldwide, with nearly 415 million diabetics at risk \cite{idf}. Early detection and treatment of DR can reduce the risk of blindness by 95\% \cite{nei}. One of the most common ways to detect diabetic eye disease is to have a specialist examine pictures of the back of the eye called fundus images and rate them on the International Clinical Diabetic Retinopathy scale \cite{opthalmology}, defined based on the type and extent of lesions (e.g. microaneurysms, hemorrhages, hard exudates) present in the image. The image is classified into one of 5 categories consisting of (1) \textit{No DR}, (2) \textit{Mild NPDR (non-proliferative DR)}, (3) \textit{Moderate NPDR}, (4) \textit{Severe NPDR}, and (5) \textit{Profilerative DR} (Figure \ref{fig:fundus_images}). Another important clinical diagnosis that can be made from the fundus image is the presence of diabetic macular edema (DME). While this work focuses only on the 5 point grading of DR, the findings should be applicable to DME diagnosis as well.

Most of the prior work on DR classification focuses on obtaining a single ground truth diagnosis for each image, and then using that for training and evaluation. Deep learning has recently been used within this setting by \citet{gulshan} who show a high sensitivity (97.5\%) and specificity (93.4\%) in the detection of referable DR (moderate or more severe DR).

\begin{figure}
\centering
\includegraphics[width=3.31in, height=2.5in]{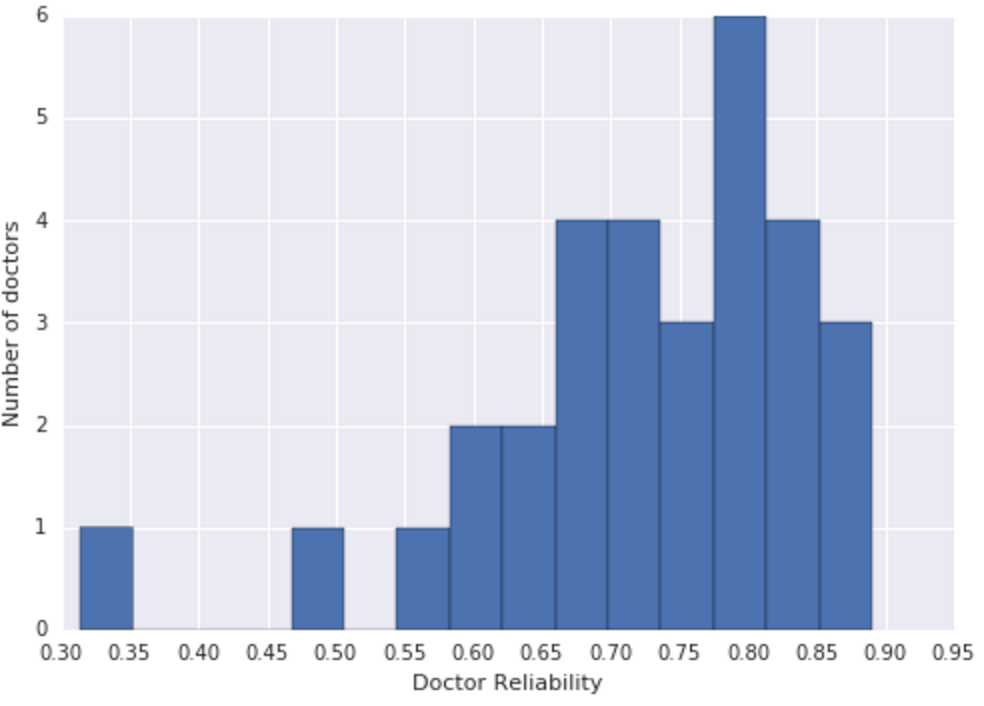}
\caption{\label{fig:reliabilities} Histogram of doctor reliabilities. \normalfont{These are calculated from the expectation-maximization algorithm in \citet{welinder} on our training data.}}
\end{figure}

In this work we explore whether, in the context of data in which every example is labeled by multiple experts, a better model can be trained by predicting the opinions of the individual experts as opposed to collapsing the many opinions into a single one. This allows us to keep the information contained in the assignment of experts to opinions, which should be valuable because experts labelling data differ from each other in skill and area of expertise (as is the case with our ophthalmologists, see Figure \ref{fig:reliabilities}). Note that we still need a single opinion on the test set to be able to evaluate the models. To that end, we use a rigorous adjudicated reference standard for evaluation, where a committee of three retinal specialists resolved disagreements by discussion until a single consensus is achieved.


\begin{figure*}
\centering
\includegraphics[scale=0.385]{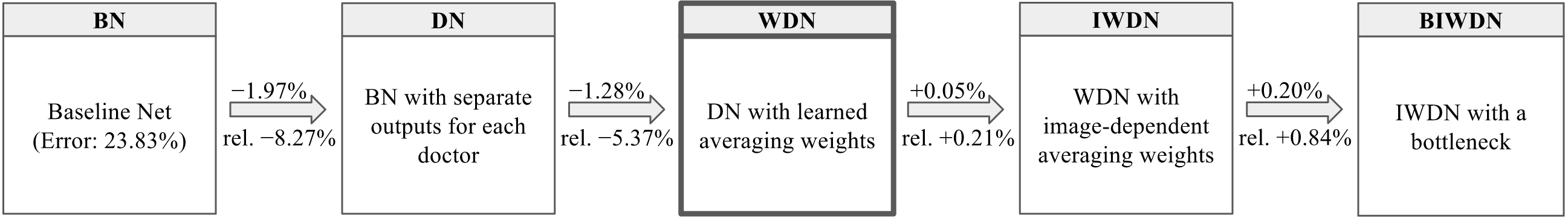}
\caption{\label{fig:schema} Description of nets used in paper. \normalfont{The baseline net has five-class classification error rate of 23.83\% on the test dataset. The numbers above arrows refer to absolute changes in test error while the numbers below arrows refer to relative changes (negative values represent improvements). WDN (highlighted) was the optimal net.}}
\end{figure*}

\section{Related Works}
Our work on learning from multiple noisy annotators relates to literature on noisy labels, crowd-sourcing, weak supervision, semi-supervised learning, item response theory, and multi-view learning.

Since the foundational work of \citet{dawid}, who model annotator accuracies with expectation-maximization (EM), and \citet{smyth}, who integrate the opinions of many experts to infer ground truth, there has a large body of work using EM approaches to estimate accurate labels for datasets annotated by multiple experts \cite{whitehill,raykar2009,raykar2012,welinder}. 

Works that use Bayesian probabilistic models for the image generation and/or annotation process include \citet{welinder2010multidimensional}; \citet{raykar2010}; \cite{wauthier}; \citet{moreno2015bayesian}. 
\citet{yan2011active}; \citet{rodrigues2014gaussian} learn from multiple annotators how to do active learning, i.e. which samples to select and which annotators to query, the latter using Gaussian processes to explicitly handle uncertainty. \citet{karger}; \cite{liu} propose message passing algorithms for crowdsourcing.

An extension in weak supervision is generalizing from noisy sources to programmatically generate labeled training sets \cite{ratner2016data}. An extension in the crowdsourcing domain is budget allocation during label sourcing \cite{chen2013optimistic} .


Previous work in biostatistics and epidemiology that estimate ground truth from multiple annotators in the absence of ground truth data are \citet{rutjes2007evaluation}; \citet{hui1980estimating}; \citet{albarqouni2016aggnet} but none of these model individual labelers as we do.

\section{Methods}
\subsection{Motivation for Model Design}
First we describe the rationale behind our proposed models. There is more information in the particular labels produced by particular doctors than is captured by simply taking the average of all the doctors who have labeled a particular image and treating this distribution as the correct answer. The amount of constraint that a training case imposes on the weights of a neural network depends on the amount of information required to specify the desired output. So if we force the network to predict what each particular doctor would say for each particular training case we should be able to get better generalization to test data, provided this does not introduce too many extra parameters. For a K-way classification task, we can replace the single softmax \cite{deeplearningbook} that is normally used by as many different K-way softmaxes as we have doctors. Of course, there will be many doctors who have not labeled a particular training image, but this is easily handled by simply not backpropagating any error from the softmaxes that are used to model those doctors. At test time we can compute the predictions of all of the modeled doctors and average them. Our belief is that forcing a neural network to model the individual doctors and then averaging at test time should give better generalization than simply training a neural network to model the average of the doctors.

We expect some doctors to be more reliable than others and we would like to give more weight to their opinions. We should this be able to do better than just averaging the opinions of the modeled doctors. After we have finished learning how to model all of the individual doctors we can learn how much to weight each modeled doctor's opinion in the averaging. This allows us to downweight the unreliable doctor models. We also expect that the doctors will have received different training and may have experienced different distributions of images so that the relative reliability of two doctors may depend on both the class of the image and on properties of the image such as the type of camera used. Our weights for averaging doctor models should therefore possibly be image-dependent.

\subsection{Model Architecture}
\begin{figure*}
\centering
\includegraphics[scale=0.3]{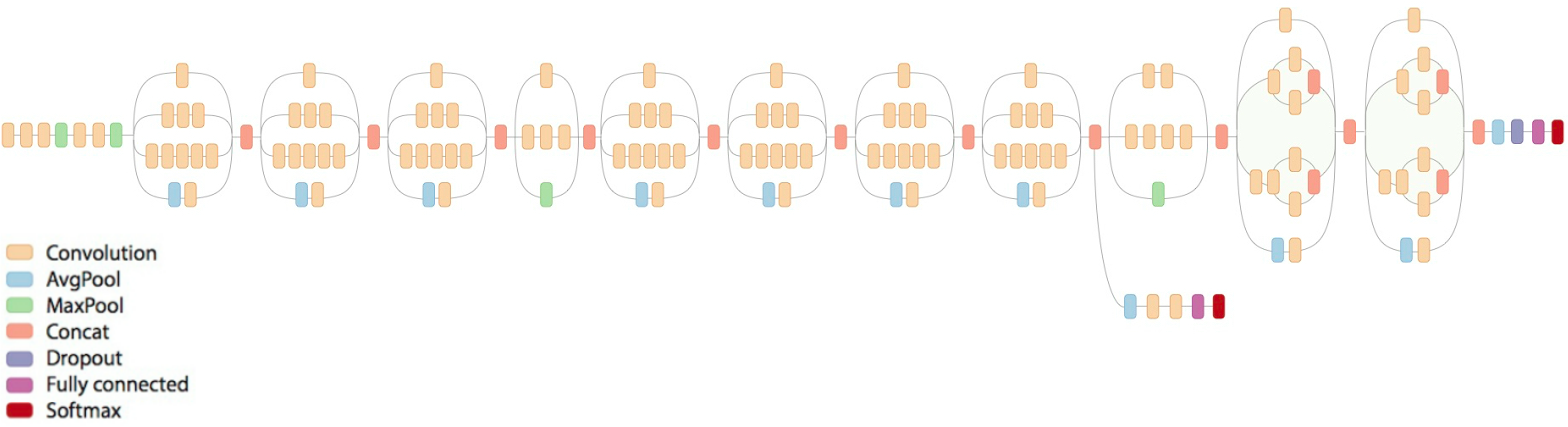}
\caption{\label{fig:Inception} Schematic diagram of Inception-v3 \cite{Inception}.}
\end{figure*}

\begin{figure*}
\centering
\includegraphics[scale=0.5]{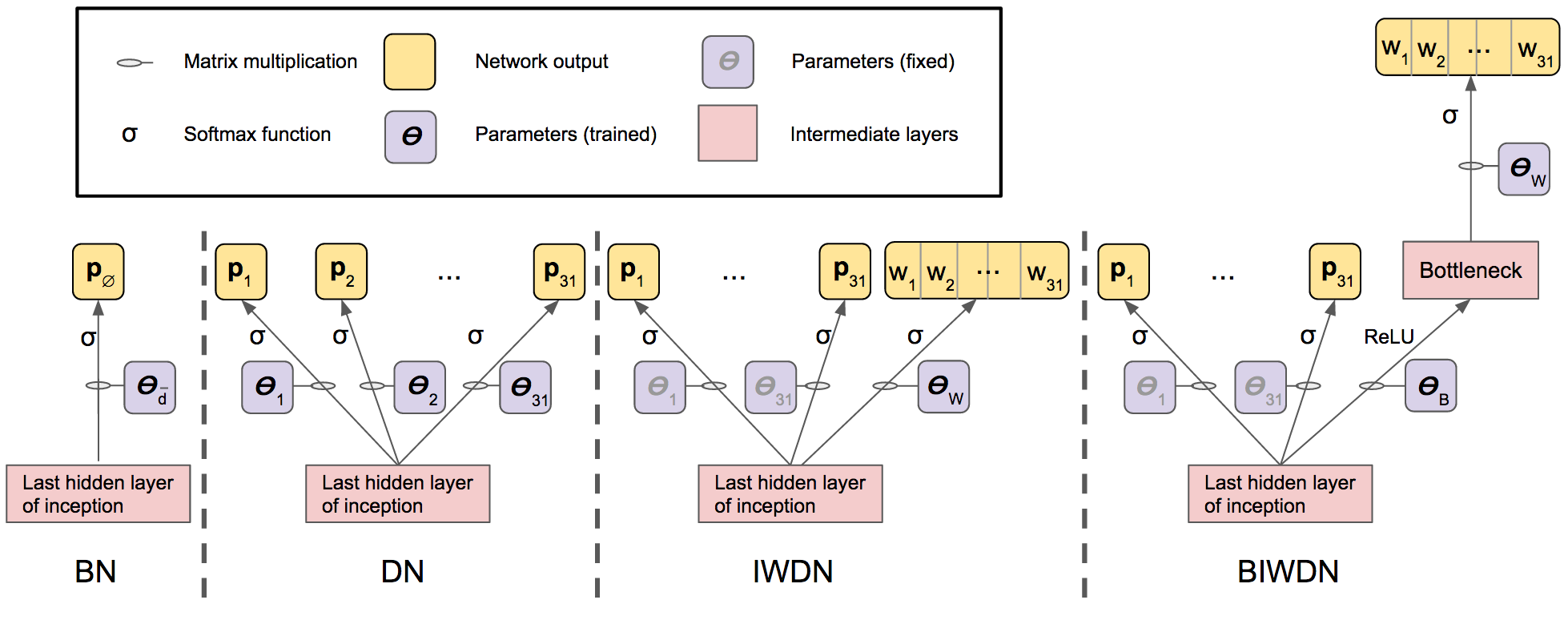}
\caption{\label{fig:computations}Schematic diagram of nets. \normalfont{These schematics show how the parameters, network outputs, and averaging weights for doctor models are connected. Table \ref{tab:computations} lists how the outputs are used in a loss function for training. In WDN (not shown in figure), the averaging logits are not connected to the last hidden layer of Inception and are just initialized from a constant vector.}}
\end{figure*}

With these intuitions in mind, we consider a sequence of models of increasing complexity for training the diabetic retinopathy classifier (Figure \ref{fig:computations}). The neural network base used in this work is the Inception-v3 architecture \cite{szegedy} (Figure \ref{fig:Inception}).
\begin{itemize}
\item \textbf{Baseline Net (BN)}: Inception-v3 trained on average opinions of doctors; a TensorFlow reimplementation of the model used in \citet{gulshan}. 
\item \textbf{Doctor Net (DN)}: BN extended to model the opinions of each of the 31 doctors.
\item \textbf{Weighted Doctor Net (WDN)}: Fixed DN with averaging weights for combining the predictions of the doctor models learned on top, one weight per doctor model.
\item \textbf{Image-specific WDN (IWDN)}: WDN with averaging weights that are learned as a function of the image.
\item \textbf{Bottlenecked IWDN (BIWDN)}: IWDN with a small bottleneck layer for learning the averaging weights.
\end{itemize}

For BN, the outputs of the last hidden layer of Inception are used to compute the logits used in the five-way softmax output layer. For DN, the opinions of each doctor are modeled using a separate softmax for each doctor, while Inception weights were shared. For evaluation, the predictions from the softmax ``doctor models'' are arithmetically averaged to give a single five-class prediction. For subsequent nets, the parameters and predictions of the DN model are frozen and only the averaging weights for the doctor models are learned. For WDN, one averaging weight per doctor is trained, shared across all images. For IWDN, these averaging weights are made image-dependent by letting them be a function of the last hidden layer of Inception. For BIWDN, a linear bottleneck layer of size three is added between the last hidden layer of Inception (of dimension 2048) and the 31-way softmax of IWDN as a precautionary measure against model overfitting. A bottleneck layer of this size reduces the number of trainable parameters ten-fold.

Rather than directly learning the averaging weight for each doctor model (B)(I)WDN, we learn averaging logits for each model that we could then pass through a softmax to produce averaging weights that are guaranteed to be positive. To train the averaging logits, we use the opinions of the doctors who actually labeled a training image to define the target output distribution for that image (Appendix B.2 discusses an alternative target). We then combine the predictions of the models of all the other doctors using the weights defined by their current averaging logits. Finally we update our parameters by backpropagating with the cross entropy loss between the target distribution and the weighted average prediction. This way all of the training cases that a doctor did {\it not} label can be used to learn the averaging logit for that doctor, and no extra data are needed beyond those used to learn the weights of DN. Moreover, if a doctor model has similar performance to other doctor models but makes very different errors it will tend to be upweighted because it will be more useful in the averaging. This upweighting of diverse doctor models would not occur if we had computed the reliabilities of the doctors separately. 

For a single image, let $I$ be the set of indices of the doctors who actually graded that image. Let the label of doctor $i\in I$ be $l_i$. For every doctor $j\in\{$1, 2, $\ldots$, 31$\}$, denote the prediction of its model $p_j$. Let $p_{\emptyset}$ be the prediction of the model of the average doctor in BN. For WDN, IWDN, and BIWDN, let $w_j$ be the averaging weight for the $j$th modeled doctor, where $\sum_j w_j=1$. Note that $p_j$ is a five-dimensional vector and $w_j$ is a scalar. The explicit inputs of the cross entropy loss being minimized during training of each model are shown in Table \ref{tab:computations} and post-Inception computations are shown schematically in Figure \ref{fig:computations}. In the case of DN, the cross entropy losses of the individual doctor models are added together to get the total loss for each training example. 

\begin{table}
\centering
\caption{\label{tab:computations} Prediction inputs to cross entropy loss for each model during training. \normalfont{The notation is given in the text. Note that the target is always $\frac{1}{|I|}\sum_{i\in I} l_i$.}}
\begin{tabular}{ccc} \toprule
Model & Training & Evaluation\\ 
\midrule
BN & \ $p_{\emptyset}$ & $p_{\emptyset}$\\\\
DN & $p_i, \forall {i\in I} $ &$\frac{1}{31}\sum_{i=1}^{31}{p_i}$\\\\
(B)(I)WDN & $\frac{\sum_{i\notin I} p_i\cdot w_i}{\sum_{i\notin I} w_i}$ &$\sum_{i=1}^{31} p_i\cdot w_i$\\
\bottomrule
\end{tabular}
\end{table}

\subsection{Summary of Procedure}

Here we summarize the entire procedure for using Weighted Doctor Net (WDN), which turns out to be the best performing model. The process is illustrated for generic labelers in Figure~\ref{fig:flowchart}.

WDN has two phases of training:
\begin{itemize}
    \item \textbf{Phase 1}: We learn a doctor model for each doctor. Each doctor model consists of the Inception-v3 base followed by a softmax output layer. The Inception-v3 is shared by all the doctor models while the output layers are unique to each doctor model. 
    \item \textbf{Phase 2}: We fix the doctor models that we learned in Phase 1 (Note this implies that the predictions made by the doctor models for any given image are also fixed.) Now we learn how to combine the predictions of the doctor models in a weighted manner. We do this by training averaging logits according to Table \ref{tab:computations} and then taking a softmax of these averaging logits to get averaging weights.
\end{itemize}

During evaluation of WDN, the prediction made for our model is a linear combination of the doctor models predictions where the coefficients are the averaging weights learned in Phase 2 of training.

Next we describe two benchmarks to compare our models against.
\subsection{Estimating Doctor Reliability with EM}

\citet{welinder} use a representative online EM algorithm to estimate abilities of multiple noisy annotators and to determine the most likely value of labels. We calculate updated labels by executing the method in \citet{welinder} on our human doctors and we use these updated labels to train BN, as a competing algorithm for our DN method. \citet{welinder} also actively select which images to label and how many labels to request based on the uncertainty of their estimated ground truth values and the desired level of confidence, and they select and prioritize which annotators to use when requesting labels. We do not use these other aspects of their algorithm because labels for all images in our dataset have already been collected. 

\subsection{Modeling Label Noise}

\citet{mnih} describe a deep neural network that learns to label road pixels in aerial images. The target labels are derived from road maps that represent roads using vectors. These vectors are converted to road pixels by using knowledge of the approximate width of the roads so the target labels are unreliable. To handle this label noise, \citet{mnih} propose a robust loss function that models asymmetric omission noise.

They assume that a true, unobserved label $\mathbf{m}$ is first generated from a $w_m\times w_m$ image patch $\mathbf{s}$ according to some distribution $p(\mathbf{m}|\mathbf{s})$, and the corrupted, observed label $\tilde{\mathbf{m}}$ is then generated from $\mathbf{m}$ according to a noise distribution $p(\tilde{\mathbf{m}}|\mathbf{m})$. The authors assume an asymmetric binary noise distribution $p(\tilde{m}_i|m_i)$ that is the same for all pixels $i$. They assume that conditioned on $\mathbf{m}$, all components of $\tilde{\mathbf{m}}$ are independent and that each $\tilde{m}_i$ is independent of all $m_{j\neq i}$. The observed label distribution is then modeled as:
$$p(\tilde{\mathbf{m}}|\mathbf{s})=\prod_{i=1}^{w_m^2}\sum_{m_i}p(\tilde{m}_i|m_i)p(m_i|\mathbf{s}).$$

For another baseline, we use a multi-class extension of their method on DN, modeling the noise distribution prior for all doctors $d$ with the parameters: $$\theta_{ll'}=p(\tilde{m}_d=l'|m_d=l)$$ where $l,l'\in\{1,2,3,4, 5\}$. We estimate $\theta_{ll'}$ using the 5$\times$5 confusion matrix between individual and average doctor opinions on training images. Treating the average doctor opinion as the true label, we convert each doctor's individual count matrix into proportions which we averaged across all doctors. We train this model by minimizing the negative log posterior, $-\log(p(\tilde{\mathbf{m}}|\mathbf{s}))$. This variant of the method by \citet{mnih} is an alternative way to improve upon DN to our proposal of learning averaging weights (WDN).

\section{Experimental Setup}

\subsection{Neural Network Training}
We train the network weights using distributed stochastic gradient descent \cite{abadi} with the Adam optimizer on mini-batches of size 8. We train using TensorFlow with 32 replicas and 17 parameter servers, with one GPU per replica. To speed up the training, we use batch normalization \cite{ioffe}, pre-initialization of our Inception network using weights from the network trained to classify objects in the ImageNet dataset \cite{russakovsky}, and the following trick: we set the learning rate on the weight matrix producing prediction logits to one-tenth of the learning rate for the other weights. We prevent overfitting using a combination of L1 and L2 penalties, dropout, and a confidence penalty \cite{pereyra}, which penalizes output distributions with low entropy. At the end of training, we use an exponentially decaying average of the recent parameters in the final model.

We tune hyperparameters and pick model checkpoints for early stopping on the validation dataset, using five-class classification error rate as the evaluation metric. The optimal values for these hyperparameters are displayed in Appendix C. Note that we tune the baseline as well to ensure that our improvements are not the result of more hyperparameter optimization. 
When evaluating on the test set we average the predictions for the horizontally and vertically flipped versions (four in total) of every image.

We also train a version of BN where the output prediction is binary instead of multi-class, as is done in \citet{gulshan}. The binary output is obtained by thresholding the five-class output at the \emph{Moderate NPDR} or above level, a commonly used threshold in clinics to define a referable eye condition. For this BN-binary network, the area under the ROC curve is used as the validation evaluation metric. 

To deal with differences in class distribution between the datasets (Table \ref{tab:dist}), we use log prior correction during evaluation. This entails adding to the prediction logits, for each class, the log of the ratio of the proportion of labels in that class in the evaluation dataset to the proportion of labels in that class in the training set. Our assumed test class distribution for computing the log prior correction is the mean distribution of all known images (those of the training and validation sets). 
So for each image under evaluation we update the prediction logit for class $c$ by adding:
\begin{align*}
& \log\bigg(\frac{q_{valid}(c)}{q_{train}(c)}\bigg)&&\textnormal{for the validation dataset, and}\\
& \log\bigg(\frac{q_{valid\cup train}(c)}{q_{train}(c)}\bigg)&&\textnormal{for the test dataset,}
\end{align*}
where $q(c)$ is the proportion of labels in that class. Applying log prior correction improves accuracy and all our reported results use it. 

\subsection{Datasets}
\label{sec:Datasets}
The training dataset consists of $126,522$ images sourced from patients presenting for diabetic retinopathy screening at sites managed by 4 different clinical partners: EyePACS, Aravind Eye Care, Sankara Nethralaya, and Narayana Nethralaya. The validation dataset consists of 7,804 images obtained from EyePACS clinics. Our test dataset consists of 3,547 images from the EyePACS-1 and Messidor-2 datasets. More details on image sourcing are in Appendix D.

Each of the images in the training and validation datasets was graded by at least one of 54 US-licensed ophthalmologist or ophthalmology trainee in their last year of residency (postgraduate year 4). For training the doctor models, we use the 30 ophthalmologists who graded at least 1,000 images were used, and we lump the remaining doctors as a single composite doctor to avoid introducing doctor-specific parameters that are constrained by less than 1,000 training cases. Meanwhile, the labels for the test set were obtained through an adjudication process: three retina specialists graded all images in the test dataset, and discussed any disagreements as a committee until consensus was reached.

We scale-normalize our images by detecting the circular fundus disk and removing the black borders around them. We use images at a resolution of 587$\times$587 pixels and we augment our training data with random perturbations to image brightness, saturation, hue, and contrast.

\subsection{Our Baseline vs Published Baseline}

This section describes the multiple ways in which our baseline differs from that of \citet{gulshan}. For these reasons, results from this paper's own BN should be used for model comparisons with DN, WN, IWDN, and BIWDN rather than numbers from \citet{gulshan}. 
\begin{itemize}
    \item Unlike in \citet{gulshan}, we remove grades of doctors who grade test set images from both training and validation sets to reduce the chance that the model is overfitting on certain experts. This handicaps our performance vis-\`{a}-vis their paper, especially because we exclude the most expert doctors (the retinal specialists) during model development, but ensures generalizability of our results.
    \item We use different datasets, and in particular our adjudicated test set has gold standard ``ground truth" labels.
    \item We train with five-class loss instead of binary loss.
    \item If a doctor grades a single image multiple times, as often occurs, \citet{gulshan} treats these as independent diagnoses while we collapse these multiple diagnoses into a distribution over classes.
    \item We employ higher resolution images (587$\times$587 pixels versus 299$\times$299) and image preprocessing and theoretical techniques unused in \citet{gulshan}.
\end{itemize}


\begin{table}
\centering
\caption{\label{tab:dist} Class distributions of datasets (as \%).}
\begin{tabular}{cccc} \toprule
Grade & Training & Validation \\  \midrule
1 & 51.03 & 72.69 \\ 
2 & 24.75 & 17.62 \\
3 & 16.81 & 7.27 \\
4 & 4.17 & 1.20 \\
5 & 3.23 & 1.21 \\\bottomrule
\end{tabular}
\end{table}

\begin{table}
\centering
\caption{\label{tab:loss} Test metrics from training with Multi-class vs Binary loss for BN.}
\begin{tabular}{ccc}
\toprule
Test Metric (\%)& Binary loss& Five-class loss\\
\midrule
Binary AUC& 95.58& 97.11\\ 
Binary Error& 11.27& 9.92\\
Spec@97$\%$Sens& 63.12& 79.60\\
\bottomrule
\end{tabular}
\end{table}

\begin{table*}
\centering
\caption{\label{tab:results} Summary of Test Results. \normalfont{All models in this table are trained with five-class loss except DN Mnih, whose loss was the negative log posterior.}}
\begin{tabular}{cccccccc}
\toprule
\multirow{2}{*}{Metric (\%)}&\multicolumn{2}{c}{BN}& \multicolumn{2}{c}{DN} & \multirow{2}{*}{WDN} &\multirow{2}{*}{IWDN}& \multirow{2}{*}{BIWDN}\\ 
&$\emptyset$ &Welinder& $\emptyset$ &Mnih & & & \\\midrule
five-class Error & 23.83 &23.74& 21.86 & 22.76& \textbf{20.58}& 20.63 & 20.83\\
Binary AUC & 97.11 & 97.00& 97.28& 97.42& \textbf{97.45} & 97.43 & 97.41\\ 
Binary Error & 9.92 & 10.12& 9.75 & 10.24& \textbf{9.07} & 9.12 & 9.23\\ 
Spec@97$\%$Sens &79.60 & 79.97& 81.81& \textbf{83.61} & 82.69& 82.46 &  82.46\\
\bottomrule
\end{tabular}
\end{table*}

\section{Summary of Results}

We run 10 replicates of each model and average the resulting metrics, which are reported in Table \ref{tab:results}. For full comparability of models we use the same 10 replicates reported for DN to serve as the fixed part of the model for training the WDN, IWDN, and BIWDN replicates.

\subsection{Training with Five-Class Loss Beats Training with Binary Loss Even on Binary Metrics}
We find that training BN with a five-class loss improves test binary AUC compared to training with a binary loss, as is done by \citet{gulshan}, even when validating the former on five-class training error instead of binary AUC (Table \ref{tab:loss}). Test binary AUC is raised by a substantial 1.53\% (97.11\% vs 95.58\%) from using five-class loss. Intuitively this fits with our thesis that generalization is improved by increasing the amount of information in the desired outputs. All results reported in Table \ref{tab:results} and subsequent sections, including for BN, are obtained from training with five-class loss.

\subsection{Averaging Modeled Doctors Beats Modeling the Average Doctor}

We see a reduction in five-class classification test error of 1.97\% (from 23.83\% to 21.86\%) from averaging modeled doctors (DN) instead of modeling the averaged doctor (BN). In comparison, using labels from the algorithm in \citet{welinder} to train BN only reduces five-class test classification error by 0.09\%. Over BN, DN also increases binary AUC by 0.17\% (97.11\% to 97.28\%), decreases binary classification error 0.17\% (9.92\% to 9.75\%), and increases specificity at 97\% sensitivity (spec@97\%sens) by 2.21\% (79.60\% to 81.81\%). Meanwhile, using labels from \citet{welinder} on BN merely increases spec@97\%sens by 0.37\% relative to vanilla BN and actually leads to slightly worse performance on binary AUC (-0.11\%) and binary error (+0.20\%). Note that the binary AUC, binary error, and spec@97\%sens metrics would be improved for all models if we were to do hyperparameter tuning and early stopping for them specifically, but we decided to do all our model selection on one metric (five-class error) both for simplicity and to simulate the decision metric required in real-life automated diagnosis systems. We see that DN is significantly better on all test metrics compared to BN trained using the labels obtained from the algorithm in \citet{welinder}.

\subsection{Learning Averaging Weights Helps}

We see a further 1.28\% decrease in five-class test error from using WDN as opposed to DN. Binary AUC increases an additional 0.17\%, binary classification error decreases another 0.68\%, and spec@97\%sens increases an extra 0.88\%, all on test data. Results from IWDN and BIWDN are slightly worse than those from WDN. We would expect a bigger improvement from WDN and potentially further improvements from training averaging logits in an image-specific way if we had doctors with more varied abilities and greater environmental differences, but on our dataset image-specific averaging logits does not help. Our extension of the competing algorithm by \citet{mnih} actually causes DN to perform worse by 0.90\% on five-class classification test error, and is more computationally costly than (B)(I)WDN. A different noise model we considered does not help either (Appendix B.3). 

\section{Conclusion}
We introduce a method to make more effective use of noisy labels when every example is labeled by a subset of a larger pool of experts. Our method learns from the identity of multiple noisy annotators by modeling them individually with a shared neural net that has separate sets of outputs for each expert, and then learning averaging weights for combining their modeled predictions. We evaluate our method on the diagnosis of diabetic retinopathy severity on the five-point scale from images of the retina. Compared to our baseline model of training on the average doctor opinion, a strategy that yielded state-of-the-art results on automated diagnosis of DR, our method can lower five-class classification test error from 23.83\% to 20.58\%. We also find that, on binary metrics, training with a five-class loss significantly beats training with a binary loss, as is done in the published baseline. We compare our method to competing algorithms by \citet{welinder}; \citet{mnih} and we show that corresponding parts of our method give superior performance to both. Our methodology is generally applicable to supervised training systems using datasets with labels from multiple annotators.

\section{Acknowledgments}
We thank Dale Webster, Lily Peng, Jonathan Krause, Arunachalam Narayanaswamy, Quoc Le, Alexey Kurakin, Anelia Angelova, Brian Cheung, David Ha, Matt Hoffman, and Justin Gilmer for helpful discussions and feedback. 
\appendix

\begin{table*}
\centering
\caption{\label{tab:hyperparam} Optimal Hyperparameters from Grid Search. \normalfont{Note that the learning rate for doctor models is one-tenth the learning rate for the rest of the network listed here. WD=weight decay, wel=welinder}}
\begin{tabular}{ccccccccc} \toprule
Hyperparameter& BN binary&BN &BN wel& DN & DN mnih&WDN &IWDN &BIWDN \\
\midrule
Learning rate& 0.0001& 0.0003 & 0.0003 & 0.001& 0.0003 & 0.03& 1$\times$10$^{-6}$& 3$\times$10$^{-7}$ \\
Dropout for Inception& 0.75 & 0.95 & 0.95 & 0.85& 0.95 & - & - & - \\
Dropout for output heads& 0.8 & 0.85 & 0.85 & 0.9& 0.9 & - & - & - \\
Entropy weight& 0.0125 & 0.025 & 0.015 & 0.0175& 0.02 & 0.0225& 0.005& 0.0125 \\
L2 WD for Inception& 0.01& 0.01 & 0.01 & 0.001& 0.004 & - & - & - \\
L1 WD for doctor models& 0.001& 0.00004& 0.0001 & 0.001& 0.01& - & - & - \\
L2 WD  for doctor models& 0.01& 0.004& 0.001 & 0.01& 0.04& - &-& -\\
L1 WD for averaging logits& - & - & -& - & -& 0.4 & 0.02& 4 \\ 
L2 WD for averaging logits& - & - & -& - & -& 15 & 0.4& 110 \\
Bottleneck size& - & - & -& - & -& -& -& 3 \\
\bottomrule
\end{tabular}
\end{table*}

\section{A. Mutual Information for Noisy Labels}

Here we compute the mutual information between a noisy MNIST label and the truth, assuming random noise, in order to estimate the number of noisily labeled training cases equivalent to one case that is known to be correctly labeled.

Empirically, $N$ perfectly labeled training cases give about the same test error as $N I_{\rm perfect}/I_{\rm noisy}$ training cases with noisy labels, where $I_{\rm noisy}$ is the mutual information per case between a noisy label and the truth and $I_{\rm perfect}$ is the corresponding mutual information for perfect labels. For ten classes, the mutual information (in nats) is $I_{\rm perfect}=2.3=-\log(0.1)$, but when a noisy label is 20\% correct on average, the mutual information is:
\begin{align*}
I_{\rm noisy} &= 0.044 \\=& -\log(0.1)-10\times0.02\times \log\bigg(\frac{0.1}{0.02}\bigg)\\
&-90\times0.1\times\frac{0.8}{9}\log\bigg(\frac{0.1}{0.1\times0.8/9}\bigg).
\end{align*}
So if the learning is making good use of the mutual information in the noisy labels we can predict that 60,000 noisy labels are worth $60,000\times0.044/2.3\approx1,148$ clean labels. In reality we needed about 1,000 clean labels to get similar results.

\section{B. Other Ideas Tested}

\subsection{B.1 Mean Class Balancing}
In addition to log prior correction of class distributions, we also attempted mean class balancing wherein samples from less frequent classes were upweighted and more frequent classes are downweighted in the cross entropy loss, in inverse proportion to their prevalence relative to the uniform distribution across classes. Explicitly, we weight each sample of class $c$ by: $$\alpha_c = \frac{\bar{q}}{q(c)}=\frac{1}{|c|q(c)}.$$ \citet{eigen} employ a similar method for computer vision tasks although they use medians instead of means. In our case, using mean class balancing lowers performance, possibly because it makes too many assumptions on the unknown test distribution, and was not employed.

\subsection{B.2. Alternative Target Distribution for Training Averaging Logits}
To train the averaging logits, we take each training case and use the opinions of the doctors who actually labeled that case to define the target output distribution. Alternatively, the target distribution can be defined as the equally weighted average of the predictions of the doctor models corresponding to the doctors who labeled that case. In the notation used in Table \ref{tab:computations}, this would be $\frac{1}{|I|}\sum_{i\in I}p_i$. We experimented with using this alternative target distribution in calculating cross entropy loss but saw inferior results.

\subsection{B.3. An Alternative Noise Model}
Because our multi-class extension of \citet{mnih} shows poor results, which we postulated may have been because it is sensitive to differences in class distributions between datasets, we considered a different noise model that makes less assumptions on the class distribution of the data. This model assumes a symmetric noise distribution that is determined by a single prior parameter. This assumes that if a label is wrong, it has equal probability of belonging to any of the other classes. However we allow this parameter to vary by doctor and we estimate it for each doctor $d$ as $$\theta_d = p(\tilde{m}_d=l|m_d=l),$$ where the real doctor reliability score is calculated from the \citet{welinder} algorithm. Unfortunately this method performs slightly worse than the 5-class variant of \citet{mnih}. Note that a number of other noise models of varying complexity can be considered as well.

\section{C. Hyperparameter Search}

Table \ref{tab:hyperparam} displays the optimal hyperparameters used in DR classification. We tuned using grid search on the following hyperparameter spaces: dropout for Inception backbone $\in \{$0.5, 0.55, 0.6, $\ldots$, 1.0$\}$, dropout for doctor models $\in \{$0.5, 0.55, 0.6, $\ldots$, 1.0$\}$, learning rate $\in \{$1$\times$10$^{-7},$  3$\times$10$^{-7}$, 1$\times$10$^{-6}$, $\ldots$, 0.03$\}$, entropy weight $\in \{$0.0, 0.0025, 0.005, $\ldots$, 0.03$\}\cup\{$0.1$\}$, weight decay for Inception $\in \{$0.000004, 0.00001, 0.00004, $\ldots$, 0.1$\}$, L1 weight decay for doctor models $\in 
\{$0.000004, 0.00001, 0.00004, $\ldots$, 0.04$\}$, L2 weight decay for doctor models $\in \{$0.00001, 0.00004, $\ldots$, 0.04$\}$, L1 weight decay for averaging logits $\in \{$0.001, 0.01, 0.02, 0.03, $\ldots$, 0.1, 0.2, 0.3, $\ldots$, 1, 2, 3, $\ldots$, 10, 100, 1000$ \}$, L2 weight decay for averaging logits $\in \{$0.001, 0.01, 0.1, 0.2, 0.3, $\ldots$,1, 5, 10, 15, 20, 30, $\ldots$, 150, 200, 300, 400, 500, 1000$\}$, and bottleneck size (for BIWDN) $\in\{$2, 3, 4, 5, 6, 7$\}$. We used a learning rate decay factor of 0.99 optimized for BN. The magnitudes of the image preprocessing perturbations were also optimized for BN. 

\section{D. Dataset Details}
Our training set consists of 119,589 of the 128,175 images used in the training set of \citet{gulshan} and 6,933 new labeled images acquired since the creation of their training dataset. The images in the training set of \citet{gulshan} that we do not use were excluded for the following reasons. (i) 4,204 images of their dataset were removed to create a separate validation dataset for experiments within the research group, (ii) 4,265 were excluded because they were deemed ungradable by every ophthalmologist that graded them. Unlike \citet{gulshan}, we do not predict image gradeability in this work and hence excluded those images. (iii) 117 of their images were excluded because they fail our image scale normalization preprocessing step. 

Our validation dataset consists of 7,963 images obtained from EyePACS clinics. These images are a random subset of the 9,963 images of the EyePACS-1 test set used in \citet{gulshan}. The remaining 2,000 images were used as part of our test set. In practice, only 7,805 of the 7,963 validation images have at least one label, since the remaining 158 images were of poor quality and considered ungradable by all ophthalmologists that labeled them.

The test set consists of 1,748 images of the Messidor-2 dataset \cite{decenciere} and 2,000 images of the EyePACS-1 test dataset used in \citet{gulshan}, as we just mentioned. 1,744 of the 1,748 images of Messidor-2 and 1,803 of the 2,000 images from EyePACS-1 were considered gradable after adjudication and were assigned labels.

\bibliographystyle{aaai}
\bibliography{bibfile}

\end{document}


\maketitle


\section{A. Mutual information for noisy labels}

Here we compute the mutual information between a noisy MNIST label and the truth, assuming random noise, in order to estimate the number of noisily labeled training cases equivalent to one case that is known to be correctly labeled.

Empirically, $N$ perfectly labeled training cases give about the same test error as $N I_{\rm perfect}/I_{\rm noisy}$ training cases with noisy labels, where $I_{\rm noisy}$ is the mutual information per case between a noisy label and the truth and $I_{\rm perfect}$ is the corresponding mutual information for perfect labels. For ten classes, the mutual information (in nats) is $I_{\rm perfect}=2.3=-log(0.1)$, but when the noisy label is 20\% correct on average, the mutual information is:
\begin{align*}
I_{\rm noisy} &= 0.044 \\=& -log(0.1)-10\times0.02\times log\bigg(\frac{0.1}{0.02}\bigg)\\
&-90\times0.1\times\frac{0.8}{9}log\bigg(\frac{0.1}{0.1\times0.8/9}\bigg).
\end{align*}
So if the learning is making good use of the mutual information in the noisy labels we can predict that 60,000 noisy labels are worth $60,000\times0.044/2.3\approx1,148$ clean labels. In reality we needed about 1,000 clean labels to get similar results.

\section{B. Other ideas tested}

\subsection{B.1 Mean Class Balancing}
In addition to log prior correction of class distributions, we also attempted mean class balancing wherein samples from less frequent classes are upweighted and more frequent classes are downweighted in the cross entropy loss, in inverse proportion to their prevalence relative to the uniform distribution across classes. Explicitly, we weight each sample of class $c$ by: $$\alpha_c = \frac{\bar{q}}{q(c)}=\frac{1}{|c|q(c)},$$ \cite{eigen} employ a similar method for computer vision tasks although they use medians instead of means. In our case, using mean class balancing lowered performance, possibly because it made too many assumptions on the hidden test distribution, and was not employed.

\subsection{B.2. Alternative target distribution for training averaging logits}
To train the averaging logits, we took each training case and use the opinions of the doctors who actually labeled that case to define the target output distribution. Alternatively, the target distribution can be defined as the equally weighted average of the predictions of the doctor models corresponding to the doctors who labeled that case. In the notation used in 
Table 1}, this would be $\frac{1}{|I|}\sum_{i\in I}p_i$. We experimented with using this alternative target distribution in calculating cross entropy loss but saw inferior results.

\subsection{B.3. A alternative noise model}
Because the multi-class extension of \cite{mnih} we tried showed poor results, which we postulated may have been because it was sensitive to differences in class distributions between datasets, we considered a different noise model that made less assumptions on the class distribution of the data. We assumed a symmetric noise distribution that is determined by a single prior parameter. This assumes that if a label is wrong, it has equal probability of belonging to any of the other classes. However we allowed this parameter to vary by doctor. For each doctor $d$ we estimated this parameter: $$\theta_d = p(\tilde{m}_d=l|m_d=l)$$ with the real doctor reliability score calculated from the \cite{welinder} algorithm. Unfortunately this method performed slightly worse than the 5-class variant of \cite{mnih}. Note that a number of other noise models of varying complexity can be considered as well.

\begin{table*}
\centering
\caption{\label{tab:hyperparam} Optimal Hyperparameters from Grid Search. \normalfont{Note that the learning rate for doctor models is one-tenth the learning rate for the rest of the network listed here. WD=weight decay, wel=welinder}}
\begin{tabular}{ccccccccc} \toprule
Hyperparameter& BN binary&BN &BN wel& DN & DN mnih&WDN &IWDN &BIWDN \\
\midrule
Learning rate& 0.0001& 0.0003 & 0.0003 & 0.001& 0.0003 & 0.03& 1$\times$10$^{-6}$& 3$\times$10$^{-7}$ \\
Dropout for Inception& 0.75 & 0.95 & 0.95 & 0.85& 0.95 & - & - & - \\
Dropout for output heads& 0.8 & 0.85 & 0.85 & 0.9& 0.9 & - & - & - \\
Entropy weight& 0.0125 & 0.025 & 0.015 & 0.0175& 0.02 & 0.0225& 0.005& 0.0125 \\
L2 WD for Inception& 0.01& 0.01 & 0.01 & 0.001& 0.004 & - & - & - \\
L1 WD for doctor models& 0.001& 0.00004& 0.0001 & 0.001& 0.01& - & - & - \\
L2 WD  for doctor models& 0.01& 0.004& 0.001 & 0.01& 0.04& - &-& -\\
L1 WD for averaging logits& - & - & -& - & -& 0.4 & 0.02& 4 \\ 
L2 WD for averaging logits& - & - & -& - & -& 15 & 0.4& 110 \\
Bottleneck size& - & - & -& - & -& -& -& 3 \\
\bottomrule
\end{tabular}
\end{table*}

\section{C: Hyperparameter Search}

Table \ref{tab:hyperparam} displays the optimal hyperparameters used computer-aided diagnosis of DR. We tuned using grid search on the following hyperparameter spaces: dropout for Inception backbone $\in \{$0.5, 0.55, 0.6, $\ldots$, 1.0$\}$, dropout for doctor models $\in \{$0.5, 0.55, 0.6, $\ldots$, 1.0$\}$, learning rate $\in \{$1$\times$10$^{-7},$  3$\times$10$^{-7}$, 1$\times$10$^{-6}$, $\ldots$, 0.03$\}$, entropy weight $\in \{$0.0, 0.0025, 0.005, $\ldots$, 0.03$\}\cup\{$0.1$\}$, weight decay for Inception $\in \{$0.000004, 0.00001, 0.00004, $\ldots$, 0.1$\}$, L1 weight decay for doctor models $\in 
\{$0.000004, 0.00001, 0.00004, $\ldots$, 0.04$\}$, L2 weight decay for doctor models $\in \{$0.00001, 0.00004, $\ldots$, 0.04$\}$, L1 weight decay for averaging logits $\in \{$0.001, 0.01, 0.02, 0.03, $\ldots$, 0.1, 0.2, 0.3, $\ldots$, 1, 2, 3, $\ldots$, 10, 100, 1000$ \}$, L2 weight decay for averaging logits $\in \{$0.001, 0.01, 0.1, 0.2, 0.3, $\ldots$,1, 5, 10, 15, 20, 30, $\ldots$, 150, 200, 300, 400, 500, 1000$\}$, and bottleneck size (for BIWDN) $\in\{$2, 3, 4, 5, 6, 7$\}$. We used a learning rate decay factor of 0.99 optimized for BN. The magnitudes of the image preprocessing perturbations were also tuned for BN. 

\section{D. Dataset Details}
119,589 of our training set images are the same as those used in the training set of \cite{gulshan} (which consists of 128,175 images). The images removed from the training dataset used by \cite{gulshan} are detailed here: (i) 4,204 out of the 128,175 were removed to create a separate validation dataset for experiments within the research group. (ii) 4,265 out of the 128,175 images were excluded since they were deemed ungradable by every ophthalmologist that graded them. Unlike \cite{gulshan}, we do not predict image gradeability in this work and hence exclude those images. (iii) $117$ out of the $128,175$ fail our image scale normalization preprocessing step and were also excluded. We also acquired 6,933 more labeled images since the creation of the training dataset in \cite{gulshan} and added them to this training set.

The validation dataset consists of 7,963 images obtained from EyePACS clinics. These images are a random subset of the 9,963 images of the EyePACS-1 test set used in \cite{gulshan}. The remaining 2,000 images were included as part of the test set in this work. In practice, only 7,805 of the 7,963 validation images have at least one label, since the remaining 158 images were of poor quality and considered ungradable by all ophthalmologists that labeled them.

The test set consists of 1,748 images of the Messidor-2 dataset \cite{decenciere} and the remaining 2,000 out of the 9,963 images of the EyePACS-1 test dataset used in \cite{gulshan}. 1,803 of the 2,000 images from the EyePACS-1 test set, and 1,744 of the 1,748 images of the Messidor-2 were considered gradable after adjudication and were assigned labels.

\bibliographystyle{aaai}
\bibliography{bibfile.bib}